\documentclass[runningheads]{llncs}
\usepackage{graphicx}
\usepackage{hyperref}
\usepackage{url}
\usepackage{multirow}
\usepackage{xcolor}
\usepackage[utf8]{inputenc}
\usepackage[russian,english]{babel}

\begin{document}
\title{A system for information extraction from scientific texts in Russian\thanks{The study was funded by RFBR according to the research project N 19-07-01134.}}
%
%
\author{Elena Bruches\inst{1,2} \and
Anastasia Mezentseva\inst{2} \and \\
Tatiana Batura\inst{1,2}\orcidID{0000-0003-4333-7888}}
\authorrunning{E. Bruches et al.}
%
\institute{A.P. Ershov Institute of Informatics Systems, Siberian Branch, Russian Academy of Sciences, Novosibirsk, Russia \\
\email{tbatura@iis.nsk.su}\\ 
\and
Novosibirsk State University, Novosibirsk, Russia \\
\email{\{e.bruches,a.mezentseva1\}@g.nsu.ru}}
\maketitle              
\begin{abstract}In this paper, we present a system for information extraction from scientific texts in the Russian language. The system performs several tasks in an end-to-end manner: term recognition, extraction of relations between terms, and term linking with entities from the knowledge base. These tasks are extremely important for information retrieval, recommendation systems, and classification. The advantage of the implemented methods is that the system does not require a large amount of labeled data, which saves time and effort for data labeling and therefore can be applied in low- and mid-resource settings. The source code is publicly available and can be used for different research purposes.

\keywords{Term extraction \and Relation extraction \and Entity linking \and Knowledge base \and Weakly supervised learning.}
\end{abstract}
\section{Introduction}
Due to the rapid growth in the number of publications of scientific articles, more and more works have recently appeared devoted to the analysis of various aspects of scientific texts. For example, the paper \cite{Head2020AugmentingSP} describes an interface that makes it easier to read scientific articles by highlighting and linking definitions, variables in formulas, etc. Authors of the work \cite{Cachola2020TLDRES} propose one of the approaches to the summarization of scientific texts. Texts of this genre contain valuable information about advanced scientific developments, however, this type of text differs from news texts, texts on social networks, etc., in their structure and content. Therefore, it is especially important to adapt and develop methods and algorithms for processing scientific texts. 

Common NLP models require a large amount of training data. However, such amounts of data are unavailable for most languages. For example, as far as we know, there are no publicly available datasets for extracting and linking entities in scientific texts in Russian. That is why we focus on the methods which do not require a large amount of labeled data: for term recognition, a weak supervision method is used; for relation extraction, we apply cross-lingual transfer learning; for entity linking, we also implement a language-agnostic method.

This paper consists of an introduction, three sections, and a conclusion. The first section provides an overview of works on the tasks of term recognition, relation extraction, and linking terms with entities from the knowledge base. The next section describes the process of preparing and annotating data. In the last section, we give a more detailed description of the system modules, present the algorithms and results of preliminary experiments.

\section{Related works}

\textbf{Terms extraction.} The goal of term extraction is to automatically extract relevant terms from a given text, where terms are sequences of tokens (usually nouns or noun groups) that define a particular concept from a field of science, technology, art, etc. There are several groups of methods for solving this task. The traditional approach solves this task in two stages: firstly, phrases which can be terms are extracted from the text, and then there is a classification step to decide whether this phrase is a term or not. Such an approach is described in \cite{bilu2020if,stankovic2016rule,zhang2018semre}. It allows control term extraction with hand-crafted rules to solve this problem more precisely. But on the other hand, the full context is rarely considered. Another approach solves this task as a sequence labeling task, for example, it is described in \cite{kucza2018term}. This group of methods takes in account the context to make use of both syntactic and semantic features. The main disadvantage of such an approach is needing quite a large amount of annotated data, as deep learning architectures are used mainly. Some researchers apply methods for topic modeling to extract terms \cite{bolshakova2013topic}. The underlying idea is that terms represent concepts related to subtopics of domain-specific texts. So revealing topics in the text collection can improve the quality of automatic term extraction.

\textbf{Relation extraction.} The relation extraction task requires the detection and classification of semantic relations between a pair of entities within a text or sentence. There are different ways to solve this task. A classical approach is to use methods based on lexico-syntactic patterns \cite{hearst-1992-automatic}. Such methods tend to have high precision and low recall since they require manual labor. To overcome this problem, nowadays different machine learning algorithms are applied  \cite{wu2019enrich,Tao2019EnhancingRE}. The distinctive feature of the method \cite{Tao2019EnhancingRE} is using a neural network for incorporation of both syntactic indicators and the entire sentences into better relation representations. However, it can be difficult to compile a complete list of such indicators. The method described in \cite{wu2019enrich} utilizes special tokens to mark two entities in the sentence which can be used in pre-trained models. This method gives good results in sentence-level relation extraction, but cannot provide information of all the entities (including multiple target entities) in the document at one time.  Since entity extraction and relation classification may benefit from having shared information, models for the joint extraction of entities and relations have recently drawn attention. In the paper \cite{ji-etal-2020-span} the authors propose a model architecture, where spans are detected and then relations between spans are classified in the end-to-end fashion. In the paper \cite{wadden-etal-2019-entity} the authors describe a method for joint entity recognition, relation extraction and event detection in a multitasking way.
This method uses separate local task-specific classifiers in the final layer, which can sometimes lead to errors because of a lack of global constraints.

\textbf{Entity linking.} The entity linking task is the task of matching an entity mentioned in a text with an entity in a structured knowledge base. Usually, this task is considered as a ranking problem and includes several subtasks. The first stage is a candidate generation for the input entity (term) from a knowledge base. For example, it can be done with string match  \cite{bunescu-pasca-2006-using}, which is rather easy to perform but doesn’t solve the problem that the same proper name may refer to more than one named entity. Another approach is dictionary  \cite{pershina-etal-2015-personalized} that helps to use taxonomy in knowledge bases but depends on their completeness. Also prior probability \cite{ganea-hofmann-2017-deep} can be used in the generation step. In the second stage, one should get embeddings for the mention with its context and for the entity. Nowadays different deep learning-based algorithms are applied for this purpose such as a self-attention mechanism \cite{luo-etal-2015-joint}, and pre-trained models \cite{yamada-etal-2016-joint}. Researchers \cite{gupta-etal-2017-entity} got mention-context vectors by using combined LSTM encoder and bag-of-mention surfaces. This approach lets model the context across datasets. However, authors didn’t include more structured knowledge to make representations semantically richer. Then there is a ranking stage to find the most relevant entity for the input term. In some research works this task is solved as a classification task, using different types of classifiers such as naive Bayes classifier \cite{varma2009}, SVM classifier \cite{zhang-etal-2010-entity}, deep neural networks \cite{HuangHJ15}.

\section{Data Description}
It is a bit complicated to find the most suitable dataset for our purposes and in the same time open-sourced. Nevertheless, the corpus of scientific papers in Russian RuSERRC\cite{bruches2020entity} solves this problem to some extent. It contains abstracts of 1.680 scientific papers on information technology in Russian, including 80 manually labeled texts with terms and relations.  We added an annotation to this corpus by linking selected terms to entities from Wikidata\footnote{\url{https://www.wikidata.org}}.

This corpus contains annotation not only of terms, but also of nested entities (entities that are inside of other entities), for example: “[self-consistent [electric field]]\foreignlanguage{russian}{([самосогласованное [электрическое поле]])}”. When annotating for entity linking task, we moved from the “largest” entity to the “smaller” nested ones, i.e. if for the very first level the entity was found in the knowledge base, then the nested entities are not annotated.

We linked terms with entities from Wikidata. They have the unique identifier prefixed with “Q”, as opposed to relations that have an identifier prefixed with “P”. Also, we did not associate terms with entities of the “Scientific article” type.
Each entity was annotated by two assessors. The measure of consistency was calculated as the ratio of the number of entities without conflict in the annotation to the total number of entities in the corpus and amounted to 82.33\%.

A total of 3386 terms were annotated in the corpus, 1337 of which were associated with entities in Wikidata. The average length of a linked entity is 1.55 tokens, the minimum length is one token, and the maximum is eight tokens.

\section{Full system architecture}
\begin{figure}[h]
    \centering
    \includegraphics[width=\textwidth]{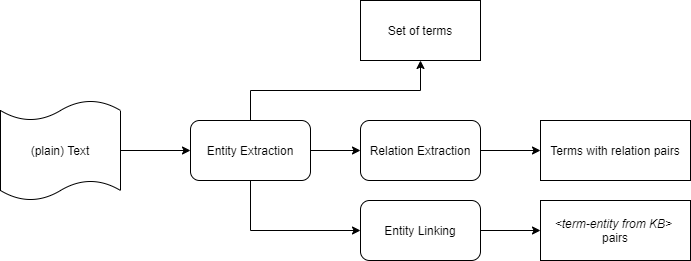}
    \caption{General architecture of the system.}
    \label{fig:architecture}
\end{figure}

We propose a framework for information extraction from scientific papers in Russian. Figure~\ref{fig:architecture} shows its general architecture.

The system consists of the following modules:

\begin{enumerate}
    \item The terms extraction module takes a raw text as input and outputs a set of terms from this text;
    \item The relation extraction module takes a text and a set of terms obtained from the previous step and outputs a set of term pairs with the specified relation between them;
    \item The entity linking module has the same input as the module for relation extraction and outputs terms and the corresponding entities from a knowledge base.
\end{enumerate}    

Below we provide a detailed description for each module.

\subsection{Entity Recognition}
\subsubsection{Method description.}

During our research, we have not found any comparative large annotated corpora for our purpose. To overcome this problem we decided to use a weak supervision method. The general idea is to train a model on data that were annotated automatically, and then annotate another data with this model, merge these two data sources and train the second model. Thus the algorithm stages are following:
\begin{enumerate}
    \item Obtaining annotated corpus for the first iteration of model training with a dictionary of terms;
    \item Training the model on this corpus;
    \item Annotating new texts and the previous ones with the trained model and the dictionary;
    \item Training the model on the extended corpus.
\end{enumerate}
	The dictionary of terms was obtained in a semi-automated way:
\begin{enumerate}
    \item We collected 2-, 3-, 4-gramms from the scientific articles, sorted them by TF-IDF value and then manually filtered them;
    \item We took titles of articles from Wikipedia, which belong to the category “Science” and then manually filtered them.
\end{enumerate}
	In such a way we obtained a dictionary with 17252 terms which is available here\footnote{\url{https://github.com/iis-research-team/ner-rc-russian}}.
	
	During all stages of our term recognition algorithm, we used the neural network architecture adopted from the BertforTokenClassification class of the Transformers library\cite{wolf-etal-2020-transformers}. Word embeddings were generated by BERT model\footnote{\url{https://huggingface.co/bert-base-multilingual-cased}}.
	
	The model takes a tokenized text (without any preprocessing) as input and outputs the sequence of labels for the corresponding tokens. Here we use three classes: “B-TERM” (for the first token in a term sequence), “I-TERM” (for the inside token in a term sequence) and “O” (for the token which doesn’t belong to a term sequence).
	
	Analysing the results, we noticed that most cases of model errors are wrong term bounds detection. To correct it we wrote some heuristics such as removing a preposition from a term if it starts the term sequence, including the next token written in English after the term sequence, etc.
	
	As a baseline for term extraction we used a dictionary-based approach - the same dictionary was used for automatic text annotation for our model. One may conclude that the dictionary-based approach gives the higher precision for partial match but still gives low recall and F1 in general. Also it can be improved by extending the list of the terms.

\subsubsection{Results.}
We tested the final model on the RuSERRC corpus, which was not used during the model training.
To evaluate the model quality, the standard classification metrics were used: precision  (P), recall (R) and F-measure (F1). Also, we considered two variants of these metrics: exact match and partial match. In exact match, only full sequences are considered to be correct. In partial match, we considered tokens that have a tag in \{“B-TERM”, “I-TERM"\} as a term. The metrics are shown in Table \ref{tab:table1}.

\begin{table}[h]
\centering
\begin{tabular}{|c|c|c|c|l|l|l|}
\hline
\multirow{2}{*}{\textbf{}} & \multicolumn{3}{c|}{\textbf{Exact match}}     & \multicolumn{3}{c|}{\textbf{Partial match}}                                                          \\ \cline{2-7} 
                           & \textit{P}    & \textit{R}    & \textit{F1}   & \multicolumn{1}{c|}{\textit{P}} & \multicolumn{1}{c|}{\textit{R}} & \multicolumn{1}{c|}{\textit{F1}} \\ \hline
Baseline                   & 0.25          & 0.17          & 0.20          & \textbf{0.82}                   & 0.34                            & 0.48                             \\ \hline
BERT                       & \textbf{0.40} & \textbf{0.31} & \textbf{0.35} & 0.77                            & \textbf{0.77}                   & \textbf{0.77}                    \\ \hline
\end{tabular}

\caption{\label{tab:table1}Term extraction metrics}
\end{table}

Relatively low metrics are largely due to the difference between the training set and gold standard annotations. As we annotated the training set with a dictionary, there were not any changes in the token sequences, while in practice the term may not include some terms, has abbreviations, reductions, etc.
Analysis of partial match metric reveals that the model is able to recognize a term but it is difficult to define the term boundaries. Considering that the task of term boundary detection is quite difficult even for humans, the obtained metrics are thought to be enough for applying this approach for solving other tasks.

\subsection{Relation extraction}
\subsubsection{Method description.} There is a lack of annotated data for the relation extraction task in Russian as well. It means that the standard training process of neural networks is difficult. To overcome this issue we applied a pre-trained multilingual model. The main idea is to finetune the pre-trained model on the data in high-resource language. Then evaluate this model on data in Russian. The hypothesis is that information from other languages encoded in the model weights helps to make predictions on data in the target language correctly.

Inspired by \cite{10.1145/3357384.3358119} we used the R-BERT architecture with pre-trained multilingual BERT model\footnote{\url{https://huggingface.co/bert-base-multilingual-cased}} and finetuned it on SciERC corpus \cite{luan-etal-2018-multi} in English, that has information about relations between scientific terms.
Since our task is to define not only the type of relation between two terms but also to define whether two terms are connected by any relation or not, we added samples without any relation to our data. To decrease the imbalance between the number of samples in classes, in the train set we added only 50\% of randomly chosen pairs of terms without relation with the distance between such terms less than 10 tokens. There were no such limitations for validation and test sets. We extracted relations only within one sentence.

For relation extraction we also implemented a pattern-based approach (baseline). The main idea is to collect lexico-syntactic patterns manually which can be used as markers for the different kinds of relations. Actually, a lot of samples don’t have an explicit marker for a particular relation - it can be found based only on words and text semantics, which is impossible to detect with hand-crafted rules.
\subsubsection{Results.} 
To evaluate the model, we used RuSERRC dataset as well. Since the sets of relation types in SciERC and RuSERRC corpora are different, we evaluated the model only on intersected relations: COMPARE, HYPONYM-OF, NO-RELATION, PART-OF, USED-FOR. The overall metrics and metrics by relations (given by the model) are shown in Table \ref{tab:table2} and Table \ref{tab:table3} correspondingly.

Zero metrics for relation COMPARE show that this relation is understood differently in these datasets. The task of relation extraction and classification is one of the most difficult tasks in NLP. Nevertheless, the obtained metrics show that this approach with some improvements can be used to solve this task without an additional set of annotated data on the target language, although we have yet to investigate this issue.

Due to the fact that we could not find an open-source good-labeled Russian dataset for our purposes we compare our results with previously published results obtained on similar datasets in English. For example, the state-of-the-art result achieved on SciERC with the SpERT (using SciBERT) method is 70.33\% for NER and 50.84\% for relation extraction. However, according to \cite{eberts2019span} the same method on the general domain dataset ACE, gives 89.28\% and 78.84\% f-scores respectively. As can be seen the f-score on the scientific dataset in English is significantly worse due to the complexity of the problem itself. Our results may also be related to insufficient data, as Russian is morphologically rich, which additionally complicates the work of the language model. We plan to study this aspect in the future. 

\begin{table}[h]
\centering
\begin{tabular}{|c|c|c|c|}
\hline
         & \textbf{Precision}     & \textbf{Recall}        & \textbf{F1}            \\ \hline
Baseline & 0.24          & 0.30          & 0.23          \\ \hline
Model    & \textbf{0.27} & \textbf{0.37} & \textbf{0.25} \\ \hline
\end{tabular}
\caption{\label{tab:table2}Relation extraction metrics}
\end{table}

\vspace{-2em}

\begin{table}[h]
\centering
\begin{tabular}{|c|c|c|c|}
\hline
\textbf{Relation} & \textbf{Precision} & \textbf{Recall} & \textbf{F1} \\ \hline
COMPARE           & 0.0              & 0.0             & 0.0         \\ \hline
HYPONYM-OF        & 0.14             & 0.42            & 0.21        \\ \hline
NO-RELATION       & 0.97             & 0.63            & 0.76        \\ \hline
PART-OF           & 0.15             & 0.15            & 0.15        \\ \hline
USED-FOR          & 0.09             & 0.69            & 0.17        \\ \hline
\end{tabular}
\caption{\label{tab:table3}Metrics by relations}
\end{table}

\vspace{-2em}

\subsection{Entity linking}
\subsubsection{Method description.}

We have implemented an algorithm for entity linking. As input data, the algorithm is given a sequence or a single token corresponding to the term. Then the stage of candidates generation, the input string undergoes preprocessing, namely lemmatization using Mystem\footnote{\url{https://yandex.ru/dev/mystem/}} and conversion to lowercase. Moreover, we take a pre-trained fastText model from Deeppavlov\footnote{\url{https://deeppavlov.ai}} and encoded input term by averaged vectors from it. Entities from Wikidata went through the same stages. Next, two main steps are performed: creating an array of candidates for linking, finding the most suitable entity in the resulting set of candidates. 

At the stage of generating candidates, the input string and its 1, 2, 3-grams are compared with the name of the entity and its synonyms. If there is a match, then the entity is added to the candidate list. In addition, if there is a "disambiguation page" in the entity's description, this candidate will be removed from the list of candidates.

For ranking candidates, we use the cosine distance between the two vectors and the threshold value for extra. The first vector is averaged for the input mention and its context of \textit{n} tokens before the term and the same number after, where \textit{n}=5. The second vector is also averaged for the name, description, and synonyms of the entity in the knowledge base. In order to take into account the number of matching tokens in candidate and mention, the calculated distances were multiplied by the weighting factor which was computed by the formula:

\begin{equation}
weight =\frac{n\_matching}{n\_all}, 
\end{equation}
\newline where $n\_matching$ is a number of shared tokens in candidate and mention; $n\_all$ is a number of all tokens in entity-mention.
The result of the algorithm is the Wikidata item identifier for the input reference. 
\subsubsection{Results.}

The algorithm was tested on a corpus with annotated scientific terms from the Data Description section. The metrics are shown in Table \ref{tab:table3}. 
We used the following metrics for evaluation: accuracy, the average number of candidates, and top-k accuracy.

\textbf{Accuracy} is the ratio of the number of correctly linked terms to the total number of terms. 
Since we managed to link not all terms in the corpus, it would be more informative to divide this metric into two: 
\textit{accuracy} and \textit{linked\_accuracy}. 

\textit{Accuracy} takes into account all entities, whereas \textit{linked\_accuracy} is calculated only on the set of terms for which the entity was found in the knowledge base in the corpus. Thus, \textit{accuracy} is computed using the formula:

\begin{equation}
accuracy = \frac{n\_correct\_entities}{ all\_entities}, 
\end{equation}
\newline where \textit{n\_correct\_entities} is the number of correctly linked terms; 
\textit{all\_entities} is the number of all terms in the corpus.

Then \textit{linked\_accuracy} is calculated by the formula:

\begin{equation}
linked\_accuracy = \frac{n\_correct\_linked\_entities}{n\_all\_linked\_entities},
\end{equation}
\newline where \textit{n\_correct\_linked\_entities} is the number of correctly linked terms among all linked terms;
\textit{n\_all\_linked\_entities} is the overall number of linked terms in the corpus.

\textbf{Average number of candidates}. We also split this metric into two:  \textit{averaged\_candidates} and \textit{linked\_averaged\_candidates}. 

\textit{Averaged\_candidates} is the average number of candidates for all entities:

\begin{equation}
averaged\_candidates = \frac{\sum_{n=1}^{n} |Candidates_{i}|}{n\_all\_entities},
\end{equation}
\newline where $Candidates_{i} $ is the set of received candidates for an entity \textit{i};
 $n\_all\_entities$ is the number of all terms in the corpus.

\textit{Linked\_averaged\_candidates} is the average number of candidates for the set of terms that managed to be linked. Denote $n\_all\_linked\_entities$ as the number of all terms in the corpus that have a link with the entity from Wikidata, and $Linked\_candidates_{i}$ as the set of generated candidates for the input term \textit{i} that was linked with Wikidata. Thus, formula for the \textit{linked\_averaged\_candidates} metric is:

\begin{equation}
linked\_averaged\_candidates = \frac{\sum_{n=1}^{n}|Linked\_candidates_{i}|}{n\_all\_linked\_entities}.
\end{equation}

\textbf{Top-k accuracy} is counted only for a set of terms in the corpus that have a relation with an entity from the knowledge base, in our context \textit{k} is equal to the number of candidates. This metric is calculated using the formula:

\begin{equation}
\textit{top-k accuracy = }
\frac{num\_correct\_sets}{n\_all\_linked\_entities},
\end{equation}
\newline where \textit{num\_correct\_sets} is the number of candidate sets for the terms which are included in the set of \textit{n\_all\_linked\_entities}, containing the true entity.

\begin{table}[]
\centering
\begin{tabular}{|l|c|c|}
\hline
\multicolumn{1}{|c|}{\textbf{Metric}} & \textbf{Baseline pipeline} & \textbf{Final pipeline} \\ \hline
Accuracy                              & \textbf{0.71}              & 0.55                    \\ \hline
Linked\_accuracy                      & 0.53                       & \textbf{0.54}           \\ \hline
Averaged\_candidates                  & 1.95                       & \textbf{10.29}          \\ \hline
Linked\_averaged\_candidates          & 2.72                       & \textbf{7.38}           \\ \hline
Tok-k accuracy                        & 0.68                       & \textbf{0.76}           \\ \hline
\end{tabular}
\caption{\label{tab:table4}Entity linking metrics}
\end{table}

A relatively high value for \textit{averaged\_candidates}, which shows the generation step works rather properly, has a bad impact on accuracy. By the way, the distinction between the value of \textit{top-k accuracy} and \textit{linked\_accuracy} is significantly large. This means that in 76\% of cases there is the top candidate in the list,  but only in 54\% of cases ranking works properly and the output entity is relevant. 

To compare results we implemented a simple algorithm for this task. It differs from the final version in two main stages. At the generation step the input string is compared with the name of the entity and its synonyms. If there is a match, then the entity is added to the candidate list. For ranking candidates, we use information about the number of links an entity has to other knowledge bases and the number of relationships of this entity with other entities. The hypothesis is that the more an entity is filled with information, the more relevant it is. Thus, the choice of entity for the input term is determined by the following formula:

\begin{equation}
\textit{linked\_entity = }
{ max (f (ent_1), …, f (ent_n))},
\end{equation}

where $n$ is the number of entities in the set of candidates,

$f (ent_i) = numL_{ent} + numR_{ent}$, 

where $numL_{ent}$ is the number of links to other knowledge bases for this entity;

$numR_{ent}$ is the number of relationships of this entity with other entities in the knowledge base.

As for metrics, only accuracy for the baseline is higher than for the final version. Probably, it is due to the significantly low value of average\_candidates. The more candidates the more difficult to find the most suitable.

As for the results of other researchers in Entity Linking in Russian, we didn’t manage to find any system or even dataset that consists of scientific papers. Nevertheless, there is a suitable dataset in English - STEM-ECR \cite{dsouza-etal-2020-stem}. Authors of this dataset evaluate several systems that work well on open domain data. Otherwise, the scores on their dataset are: exact title match heuristic at 37.8\% accuracy, and the best is for Babelfy \cite{moro-etal-2014-entity} to DBpedia at 52.6\%.

In the future, we plan to implement an approach to identify the semantic similarity of entities using a classifier based on the Siamese network. Moreover, alternative names or synonyms will be used for expanding the list of candidates.

\section{Conclusion}
In this paper, we presented a system for information extraction from scientific texts in Russian. It consists of three modules. The first module recognizes terms, the second one extracts the relations between the terms found in the previous step, and the third one links terms with entities from the knowledge base. As a result, information from the input text is extracted in a structured form. The experiments were carried out with texts from the information technology domain and, presumably, can be easily adapted to other subject areas. However, in order to draw conclusions regarding the transfer to other domains, in the future, we plan to conduct an additional series of experiments. 

Our research is publicly available  at \url{https://github.com/iis-research-team/terminator}. The results of this study can be useful in the development of systems for unstructured data analysis and expert systems in scientific organizations and higher education institutions.

\bibliographystyle{splncs04}
\bibliography{references}

\end{document}